# TAGROUTER: Learning Route to LLMs through Tags for Open-Domain Text Generation Tasks

Zhou Chen[1*], Zhiqiang Wei[2], Yuqi Bai[1†], Xue Xiong[2†], Jianmin Wu[2†]
[1]Tsinghua University, [2]AI Cloud Group, Baidu Inc.

chenz22@mails.tsinghua.edu.cn, weizhiqiang@baidu.com
YuqiBai@mail.tsinghua.edu.cn, xiongxue@baidu.com, wujianmin@baidu.com

## Abstract

Model routing allocates queries to the suitable model, improving system performance while reducing costs. However, existing routing methods face practical limitations that hinder scalability in large-scale applications and struggle to keep up with the rapid growth of the large language model (LLM) ecosystem. To tackle these challenges, we propose TAGROUTER, a training-free model routing method designed to optimize the synergy among multiple LLMs for open-domain text generation tasks. Experimental results demonstrate that TAGROUTER outperforms 13 baseline methods, increasing the accept rate of system by 6.15% and reducing costs by 17.20%, achieving optimal cost-efficiency. Our findings provides the LLM community with an efficient and scalable solution for model ensembling, offering users an evolvable "super model."

## 1 Introduction

Large Language Models (LLMs) have revolutionized the landscape of Natural Language Processing (NLP) by transforming a wide array of NLP tasks into text generation task, outperforming specialized models in various domains (Liang et al., 2024; Chen et al., 2025c). The remarkable capabilities of LLMs have attracted significant investments from both academia and industry, accelerating their advancement and widespread application. In 2024, significant advancements were marked by the release of GPT-4 (OpenAI, 2024) of OpenAI, ERNIE 4.0 (Baidu, 2024) of Baidu, and Qwen2.5 (Qwen et al., 2025) of Alibaba. Presently, the Hugging Face platform hosts over 170,000 models employed in text generation, each varying in architecture, size, training data, and method, leading to a diverse range of capabilities (Raiaan et al., 2024).

*Work done during internship at Baidu AI Cloud Group
†Corresponding author

| Dataset | Win (%) | Tie (%) | Loss (%) |
|---|---|---|---|
| Alpaca | 12.08 | 16.21 | 71.71 |
| Dolly | 17.20 | 20.42 | 62.38 |
| BCUQ | 21.39 | 39.19 | 39.42 |

Table 1: Win, Tie, and Loss rates of a smaller LLM (ERNIE-Speed-8K) compared to a larger LLM (ERNIE-3.5-8K) on the three datasets. Datasets and evaluation details are introduced in Sec. 2.3 and Sec. 3.3. We can see the smaller LLM demonstrates comparable (Tie (%)) or even superior (Win (%)) performance to the larger model on specific samples across three datasets.

The selection of LLMs for specific tasks and scenarios is often guided by their performance on relevant evaluation benchmarks (Chen et al., 2025b). Generally, models with larger parameter sizes tend to achieve higher scores on these benchmarks (Kaplan et al., 2020). However, these top scores typically represent the average performance across the benchmarks. Given the diverse capabilities of models, which allow them to demonstrate varying strengths across different queries (Tab. 1), it is crucial to evaluate their performance at the sample level (Jiang et al., 2023).

As the LLM community advances, the integration of diverse models through model routing promises to enhance capabilities of model system and reduce the reliance on larger LLMs (Patil et al., 2024; Srivatsa et al., 2024). The model routing system automates the selection of the optimal candidate model in model system for each query by capturing its semantic features and generating responses (Sakota et al., 2024). The routing system streamlines user interaction by automatically selecting the suitable model, minimizing the complexity and effort involved in searching and testing multiple different models (Ding et al., 2024).

Most studies conceptualize model routing as a multi-label classification problem (Lu et al., 2024a), yet there remains substantial room for improve-ment. Methods requiring the multiple call of candidate models can lead to increased latency and higher system costs (Jiang et al., 2023; Yue et al., 2024; Chen et al., 2023). Other methods fail to manage usage costs effectively, limiting their feasibility for large-scale deployment (Lu et al., 2024d; Tekin et al., 2024; Lu et al., 2024b). Methods like Leviathan et al. (2023); Sun et al. (2024); Ramírez et al. (2024) require access to logits during inference, complicating the routing control for proprietary models. Furthermore, task-specific methods and those requiring specially designed loss functions present scalability challenges for open-domain tasks (Aggarwal et al., 2023; Mohammadshahi et al., 2024; Nguyen et al., 2024). Although some methods address certain shortcomings, they require retraining whenever there are changes in the candidate models, which reduces their adaptability in the fast-evolving LLM ecosystem (Hari and Thomson, 2023; Sakota et al., 2024; Liu et al., 2024). Moreover, some methods only support routing between two models with different parameter scales, which limits their scalability for tasks involving multiple models or models with minimal differences in capabilities (Lee et al., 2024; Ong et al., 2024; Ding et al., 2024).

This work introduces TAGROUTER, a practical routing method for LLMs that leverages self-aware tags. TAGROUTER captures key semantic features of user queries and controls the behavior of multiple models. It seamlessly ensembles models in a training-free manner, while controlling costs and meeting the requirements of open-domain text generation tasks. By leveraging these capabilities, TAGROUTER improves the efficiency of the increasingly complex model ecosystem, offering users a evolvable "super model." The contributions of this work are as follows:

- We developed TAGROUTER, a novel model routing method that enhances model system performance by ensembling multiple LLMs. TAGROUTER outperforms 13 baseline methods in open-domain text generation tasks, providing a more cost-efficient and scalable solution for model routing.

- TAGROUTER is the first routing method with six features: training-free, support for open-domain text generation tasks, multi-candidate model routing, proprietary models, cost control, and no repeated model calls. These features improve routing system practicality and offer new perspectives for future research.

- In addition to TAGROUTER, we proposed three tag-based routing methods that surpassed existing routing methods. These tag-based methods introduced a novel framework for model routing, contributing to the advancement of research in this area.

## 2 Preliminaries

### 2.1 Model Routing

Model routing can be classified into three types based on the sequence in which the *routing system assigns the query* and the *candidate model performs inference*.

**Routing after inference** involves selecting a suitable model based on the quality of generated responses (Aggarwal et al., 2023; Tekin et al., 2024; Ramírez et al., 2024). FrugalGPT (Chen et al., 2023) sorts model parameters by size and perform inference iteratively until the response meets a predefined quality threshold. LLM-Blender (Jiang et al., 2023) ranks responses via PairRanker and integrates the top three using GenFuser. Yue et al. (2024) argue against using larger LLMs when smaller ones yield consistently high-quality responses. These methods ensure precise routing but increases latency and system costs.

**Routing during inference** involves routing decisions made during the decoding process of model inference (Leviathan et al., 2023; Sun et al., 2024). BiLD (Kim et al., 2024) primarily uses a smaller model and resorts to a larger one when necessary. Li et al. (2024) combine outputs from various models to address data poisoning and privacy issues. These methods boost efficiency but struggles with heterogeneous architectures and scalability.

**Routing before inference** refers to the routing occurring before any model response generation. FORC (Sakota et al., 2024) embeds a model identifier into the input, predicting performance via DistilBERT. RouteLLM (Ong et al., 2024) sorts models into tiers and simplifies selection to a binary classification task. RouterBench (Hu et al., 2024) uses KNN and MLP for routing decisions. While these methods help reduce latency and costs, they may compromise routing system performance and require frequent updates as the models evolve.

## 2.2 Problem Setup

This work aims to address the challenge of assigning different queries to the most suitable LLMs within a model system, thereby enabling the performance of system to exceed that of any individual model. Let $\mathcal{M} = \{M_1, \ldots, M_i\}$ denote the model system, and $\mathcal{Q} = \{q_1, \ldots, q_n\}$ denote the set of queries. The objective is to assign each query $q \in \mathcal{Q}$ to a model $M \in \mathcal{M}$ in order to maximize the collective performance of the model system.

We propose a tag-based routing method for model routing. We believe that using a tag generation model $\mathcal{T}$ to generate a set of tags $\mathcal{T}(q)$ for each query $q$ can improve the routing performance. The routing decision is then determined by the following function:

$$M^*(q) = \text{argmax}_{M \in \mathcal{M}} f(\mathcal{T}(q), M),$$

where $f$ quantifies the alignment between the generated tags $\mathcal{T}(q)$ and the capabilities of each model $M$, producing a utility score that predicts the efficacy of the model $M$ in handling the query $q$. The model $M^*(q)$ with the highest utility score is selected to response the query.

## 2.3 BCUQ: A Real-World Benchmark

This work employs the Baidu AI Cloud User Queries (BCUQ) dataset as a benchmark for evaluating open-domain text generation tasks. The BCUQ dataset contains 95,559 user query logs from the ERNIE Bot platform on Baidu AI Cloud, representing user needs and behavioral patterns in real-world. It encompasses eight types of tasks, including classification and brainstorming (Fig. 4). To ensure user privacy, all experiments involving user data were conducted in a secure cloud environment on Baidu AI Cloud.

## 3 TAGROUTER

### 3.1 Overview

TAGROUTER consists of three modules: TAGGENERATOR, TAGSCORER, and TAGDECIDER. These modules are designed for practical applicability. The TAGGENERATOR is query-agnostic and does not require retraining. The TAGSCORER stores a key-value mapping derived from the performance of each candidate model in handling different tags, evaluated on the dataset. The TAGDECIDER provides default threshold values for cost-efficient routing and an optimization method tailored to specific scenarios, eliminating the need for manual threshold tuning. This design enables a lightweight, training-free routing process and facilitates the seamless extension of candidate models.

### 3.2 TAGGENERATOR

We train the TAGGENERATOR[1] to generate a set of tags $\mathcal{T}(q) = \{t_1, t_2, \ldots, t_j\}$ for a given query $q$, where $t_j$ represents a specific semantic feature or attribute of the query, and $j$ is the index of the tags associated with the query. These tags are crucial for routing queries to the most suitable model based on their respective capabilities.

**Tagging.** Unlike fixed tag sets, we utilize an open-tagging approach. For each query $q$, we prompting ERNIE-4.0-Turbo-8K (denoted as EB4.0) to generate tags $\mathcal{T}(q)$ (Appx. H). This approach ensures that the generated tags are flexible and diverse, helping to capture the varied user intents that may not be covered by predefined tag sets. As a result, we generate a raw set of 14,352 unique tags over the BCUQ dataset.

**Normalization.** To improve robustness and reduce noise in the generated tags, we apply the following normalization techniques: *(i)* **Frequency Filtering:** Discard rare tags appearing fewer than five times and focus on more frequent and reliable tags. *(ii)* **Rule Aggregation:** Replace special characters with spaces and capitalize the first letter of each word to standardize the tag format. *(iii)* **Semantic Aggregation**: We use PhraseBERT (Wang et al., 2021a) embeddings to represent each tag and apply DBSCAN clustering to group similar tags. Through an iterative merging process (Alg. 1), tags are consolidated into broader categories, ensuring each cluster contains at least two tags. This approach improves the model ability to distinguish related but distinct tags, simplifying the tag structure while preserving essential semantic information. After the tag normalization, we obtain a refined tag set containing 1,601 unique tags.

**Training the TAGGENERATOR.** We train the TAGGENERATOR using knowledge distillation. The training dataset is defined as:

$$\mathcal{D} = \{(q, \mathcal{T}(q)) \mid q \in \mathcal{Q}\},$$

where each sample $(q, \mathcal{T}(q))$ consists of a query $q$ and its corresponding set of tags $\mathcal{T}(q)$.

Firstly, we apply the Hybrid Weight-Based Data Sampling algorithm (Alg. 2) to sample the training

---
[1] https://huggingface.co/itpossible/TagGenerator

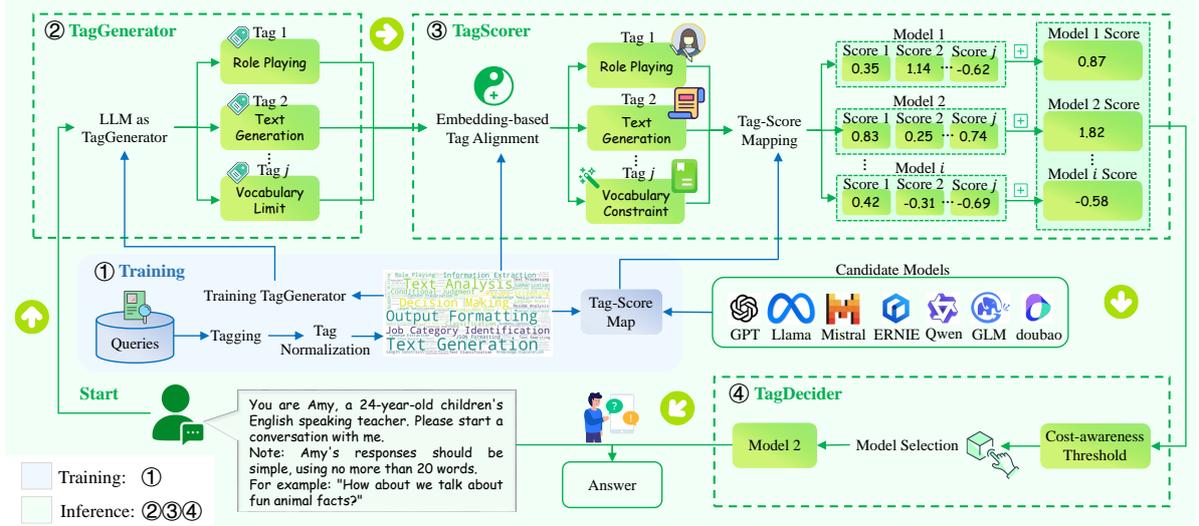

Figure 1: Overview of TAGROUTER. The training phase is represented in blue, and the inference phase in green. TAGROUTER consists of three modules: TAGGENERATOR, TAGSCORER, and TAGDECIDER, which are invoked sequentially. First, TAGGENERATOR generates fine-grained tags for each query. Next, TAGSCORER evaluates the performance of different models on the query by computing scores based on these tags. Finally, TAGDECIDER selects the appropriate model for inference, considering both the computed scores and a cost-awareness threshold.

dataset $\mathcal{D}$, prioritizing rare but significant tags. The sampled data is then used to train TAGGENERATOR through instruction tuning on a smaller LLM like Qwen2.5-0.5B.

### 3.3 TAGSCORER

TAGSCORER evaluates the performance of each candidate model in model system $\mathcal{M}$ in handling queries. For a given query $q$ and its corresponding tags $\mathcal{T}(q)$, TAGSCORER computes a score for each model $M_i \in \mathcal{M}$, reflecting the model ability to interpret the semantic of the query.

**Tag Alignment.** To address mismatches between generated tags and the tag set, we introduce an embedding-based tag mapping method. We use PhraseBERT embeddings (Wang et al., 2021a) to represent each tag $t \in \mathcal{T}(q)$ and calculate the cosine similarity between a generated tag and each tag in the tag set. The most similar tag is then selected to map generated tags into a unified tag space, enhancing consistency.

**Tag-Score Mapping.** We define the reference model $M_{\text{LLM}}$ as the model with the largest parameter size in the model system $\mathcal{M}$, which serves as the baseline for pairwise comparisons and performance evaluation. For each model $M_i$ and tag $t$, we calculate the performance score $\text{score}(M_i, t)$, which is defined as:

$$\text{score}(M_i, t) = w_t \cdot \sum_{r \in \{\text{win,tie,loss}\}} \text{count}_{t,M_i}(r) \cdot s_r,$$

where $\text{count}_{t,M_i}(r)$ denotes the frequency of result $r \in \{\text{win}, \text{tie}, \text{loss}\}$ for tag $t$ and model $M_i$, and $s_r$ represents the score associated with result $r$. Specifically, $s_{\text{win}}$, $s_{\text{tie}}$, and $s_{\text{loss}}$ are the score weights for wins, ties, and losses, respectively. The result $r$ is determined through pairwise comparisons, prompted by EB4.0. The weight $w_t$ reflects the confidence in tag $t$, which is defined as:

$$w_t = \frac{1 - \exp(-\text{count}_t)}{\sum_{t' \in \mathcal{T}} \text{count}_{t'}},$$

where $\text{count}_t$ is the frequency of tag $t$ in the training dataset $\mathcal{D}$.

Thus, $\text{score}(M_i, t)$ quantifies the relative performance of model $M_i$ on tag $t$, normalized by the tag frequency. This adjustment ensures that both commonly occurring tags and those with low frequency but high consistency in comparison results have a more significant impact on the selection of the optimal model $M^*(q)$.

### 3.4 TAGDECIDER

The TAGDECIDER module selects the optimal model $M^*(q)$ for each query $q \in \mathcal{Q}$ based on the scores generated by the TAGSCORER. The set of optimal models for all queries is denoted as $\mathcal{M}^* = \{M^*(q) \mid q \in \mathcal{Q}\}$. For each query $q$, the optimal model $M^*(q)$ is selected as:

$$M^*(q) = \text{argmax}_{M \in \mathcal{M}} \sum_{t \in \mathcal{T}(q)} \text{score}(M, t),$$

where $\text{score}(M, t)$ represents the performance score of model $M$ with respect to tag $t$. This function ensures that query $q$ is routed to the model $M^*(q)$ that maximizes the cumulative alignment between the model and the tags that best characterize the semantic features of query.

In real-world applications, model selection often involves considering cost. This cost is managed by defining a cost-awareness threshold $\theta$. When a query $q$ is routed to $M_{\text{LLM}}$, the score difference $\Delta_q$ between the smaller model $M_{\text{SLM}}(q)$ and $M_{\text{LLM}}(q)$ is computed as follows:

$$\Delta_q = \sum_{t \in \mathcal{T}(q)} \text{score}(M_{\text{SLM}}(q), t) - \text{score}(M_{\text{LLM}}(q), t)$$

where $\mathcal{T}(q)$ denotes the set of tags associated with query $q$, and $\text{score}(M, t)$ is the performance score of model $M$ on tag $t$.

If $\Delta_q < \theta$, the query is routed to $M_{\text{LLM}}(q)$ ($M^*(q) \to M_{\text{LLM}}(q)$); otherwise, it is routed to $M_{\text{SLM}}(q)$ ($M^*(q) \to M_{\text{SLM}}(q)$).

The routing method is expected to perform optimally when $\theta = 0$. Using $\theta = 0$ as a baseline, lowering $\theta$ shifts the focus of system toward cost, increasing the likelihood of routing queries to lower-cost models. By dynamically adjusting $\theta$, the cost of system can be controlled while maintaining performance that surpasses individual models.

## 4 Evaluation Metrics

**Accept Rate (AR)** quantifies the proportion of queries $q \in \mathcal{Q}$ for which the responses generated by the optimal model $M^*(q)$ meet the expected outcomes (surpassing those generated by $M_{\text{LLM}}$), including both "win" and "tie" responses. AR is defined as:

$$\text{AR} = \frac{\sum_{q \in \mathcal{Q}} \text{count}_{M^*(q)}(\{\text{win}, \text{tie}\})}{|\mathcal{Q}|}$$

where $\text{count}_{M^*(q)}(\{\text{win}, \text{tie}\})$ represents the number of responses generated by model $M^*(q)$ that are classified as either "win" or "tie".

**GPT-Rank (Rank)** denotes the average ranking of model $\mathcal{M}^*$ on dataset $\mathcal{Q}$. A value of 1 indicates that $\mathcal{M}^*$ achieves the highest performance on $\mathcal{Q}$.

**Area Under Curve (AUC)** evaluates the performance of the model system by computing the area under the curve defined by the routing ratio $\rho$ to $M_{\max}$ along the x-axis and the corresponding AR values along the y-axis. The AUC is defined as:

$$\text{AUC} = \int_0^1 \text{AR}(\rho) \, d\rho.$$

**Partial Area Under Curve (PAUC)** measures the performance of model system in regions where the AR surpasses that of $M_{\text{LLM}}$. Specifically, PAUC represents the area under the AUC curve where $\text{AR}(\rho) > \text{AR}_{M_{\text{LLM}}}$, with $\text{AR}_{M_{\text{LLM}}}$ denoting the AR achieved by always routing to $M_{\text{LLM}}$. The PAUC is defined as:

$$\text{PAUC} = \int_{\text{AR}(\rho) > \text{AR}_{M_{\text{LLM}}}} \text{AR}(\rho) \, d\rho.$$

A higher PAUC score indicates that the routing system more effectively selects models $M^*(q)$ that outperform $M_{\text{LLM}}(q)$. Therefore, PAUC serves as a key metric for evaluating the ability of the routing system $f(\mathcal{T}(q), M)$ to enable the performance of the model system $\mathcal{M}$ to surpass that of $M_{\text{LLM}}$.

## 5 Experiments

### 5.1 Experimental Settings

**Training and Inference.** We trained TAGGENERATOR using eight A100 80GB GPUs on a sampled version of BCUQ dataset (sampling procedure described in Alg. 2). To identify the optimal base model, we explored several model series with different parameter scales, as detailed in Tab. 9. Qwen2.5-0.5B was chosen as the base model for training TAGGENERATOR, owing to its superior balance between performance and computational efficiency. All validation experiments were conducted on two A100 80GB GPUs.

**Candidate Models.** The candidate models include ERNIE-3.5-8K (denoted as EB3.5) and ERNIE-Speed-8K (denoted as EBspeed), both developed by Baidu. To assess the training-free adaptability of TAGROUTER, we incorporated three additional models: EBspeedX (a variant of EBspeed), GLM4-9B, and Qwen2.5-7B. Among these five models, EB3.5 has the largest parameter size, highest performance, and cost. Therefore, we designate EB3.5 as $M_{\text{LLM}}$, and the others as $M_{\text{SLM}}$. The goal of TAGROUTER is to optimize the model system to outperform EB3.5 while reducing costs.

**Baselines.** We established the following models and baseline methods for comparison: *(i)* **Individual Model:** Evaluation of individual models on the benchmark dataset. *(ii)* **Existing Routing Methods:** Implementation and reproduction of ten routing methods, with hyperparameter tuning to select the best-performing configurations (Appx. B.2). *(iii)* **Tag-based Methods:** By converting the input from query $q$ to the tags $\mathcal{T}(q)$, we retrain the top three existing routing methods. Specifically, the training process is represented by $f : \mathcal{T}(q) \to M^*(q)$.

| Category | Method | Performance at Max AR | | | | AUC(%)↑ | PAUC(%)↑ |
|---|---|---|---|---|---|---|---|
| | | AR(%)↑ | Uplift(%)↑ | Cost↓ | Rank↓ | | |
| Individual LLM | EBspeed | 59.78 | -24.1 | 2.01 | 1.400 | - | 0 |
| | EB3.5 | 78.76 | 0 | 13.49 | 1.212 | - | 0 |
| Existing Routing Methods | FrugalGPT (Chen et al., 2023) | 78.88 | 0.15 | 13.24 | 1.211 | 70.11 | 0.01 |
| | PairRanker (Jiang et al., 2023) | 78.76 | 0 | 13.49 | 1.212 | 72.17 | 0 |
| | Blending (Lu et al., 2024d) | 78.76 | 0 | 13.49 | 1.212 | 69.22 | 0 |
| | RouteLLM$^{SWR}$ (Ong et al., 2024) | 78.76 | 0 | 13.49 | 1.212 | 70.88 | 0 |
| | RouteLLM$^{BERT}$ (Ong et al., 2024) | 78.76 | 0 | 13.43 | 1.212 | 71.35 | 0 |
| | RouteLLM$^{LLM}$ (Ong et al., 2024) | 78.76 | 0 | 13.49 | 1.212 | 73.02 | 0 |
| | RouteLLM$^{MF}$ (Ong et al., 2024) | 80.34 | 2.01 | 11.82 | 1.197 | 73.94 | 0.12 |
| | RouterBench$^{MLP}$ (Hu et al., 2024) | 78.88 | 0.15 | 13.40 | 1.211 | 73.58 | 0.01 |
| | RouterBench$^{KNN}$ (Hu et al., 2024) | 80.45 | 2.15 | 11.77 | 1.196 | 75.15 | 0.40 |
| | FORC (Sakota et al., 2024) | 81.80 | 3.86 | 11.81 | 1.182 | 75.73 | 0.76 |
| Tag-based Methods (ours) | RouteLLM$^{MF}$ w/ TAGGENERATOR | 82.02 | 4.14 | 11.66 | 1.180 | 76.08 | 0.76 |
| | RouterBench$^{KNN}$ w/ TAGGENERATOR | 81.57 | 3.57 | 11.76 | 1.184 | 74.48 | 0.98 |
| | FORC w/ TAGGENERATOR | 81.91 | 4.00 | 11.79 | 1.181 | 75.97 | 0.59 |
| | **TAGROUTER** | **83.60** | **6.15** | **11.17** | **1.164** | **76.10** | **1.46** |

Table 2: Performance of TAGROUTER and baselines on BCUQ dataset. **Bold** numbers indicate the best results among all routing methods, and the second-best results are underlined. TAGROUTER outperforms all baselines.

### 5.2 Experimental Results

#### 5.2.1 Performance on BCUQ

Tab. 2 presents the performance comparison of TAGROUTER and baselines on the BCUQ dataset, with EBspeed and EB3.5 as candidate models.

**Model routing enhances performance of model system.** Most of routing methods outperform EB3.5 in both AR and rank metrics, underscoring the efficacy of model routing in dynamically selecting a suitable model based on query characteristics. By ensembling multiple models and leveraging their complementary strengths, it enhances system efficiency and performance.

**TAGGENERATOR improves routing performance by encoding query semantics into informative tags.** Compared to existing routing methods that rely on raw queries, tag-based methods demonstrate significant improvements in both AR and rank metrics. For example, RouteLLM$^{MF}$ with TAGGENERATOR outperforms the standard RouteLLM$^{MF}$. These results suggest that tags capture key semantic features effectively while filtering irrelevant information, thereby enhancing both the generalization capability and decision-making efficiency of routing systems.

**TAGROUTER achieves SOTA performance.** *(i)* TAGROUTER consistently outperforms individual LLMs, exising routing methods, and other tag-based methods across both AR and rank metrics, demonstrating its superior ability to allocate queries to appropriate models. It boosts the AR by 6.15% while reducing costs by 17.20%, showcasing optimal cost-efficiency. *(ii)* TAGROUTER attains the highest AUC score, indicating its robustness in selecting optimal candidate models under varying cost constraints. *(iii)* TAGROUTER achieves the highest PAUC score, indicating its superior competitive edge in ensuring the performance of model system surpasses that of any individual candidate model like $M_{LLM}$.

#### 5.2.2 Performance Across Different Tasks

Fig. 2 presents the comparison of TAGROUTER and the top three ranking exiting routing methods across eight task categories in the BCUQ, with EBspeed and EB3.5 as candidate models.

**LLMs exhibit distinct strengths and limitations across different task categories.** In seven task categories, EB3.5 achieves a higher AR score than EBspeed. However, in the summarization task, EBspeed surpasses EB3.5 in AR metric. This suggests despite its larger parameter size, EB3.5 does not consistently outperform the smaller EBspeed across all task categories. Therefore, the model routing, which assigns queries to suitable candidate model rather than defaulting to the largest LLM $M_{LLM}$, emerges as a cost-efficient method.

**The effectiveness of routing methods varies between task categories.** In tasks such as brainstroming and content creation, the four routing methods significantly outperform *random routing*. However, in close QA and open QA tasks, their per-formance remains comparable to that of *random routing*. This could be attributed to the structured nature of QA queries, which often follow similar patterns, making it difficult for the routing system to distinguish fine-grained variations within QA tasks based solely on semantic cues.

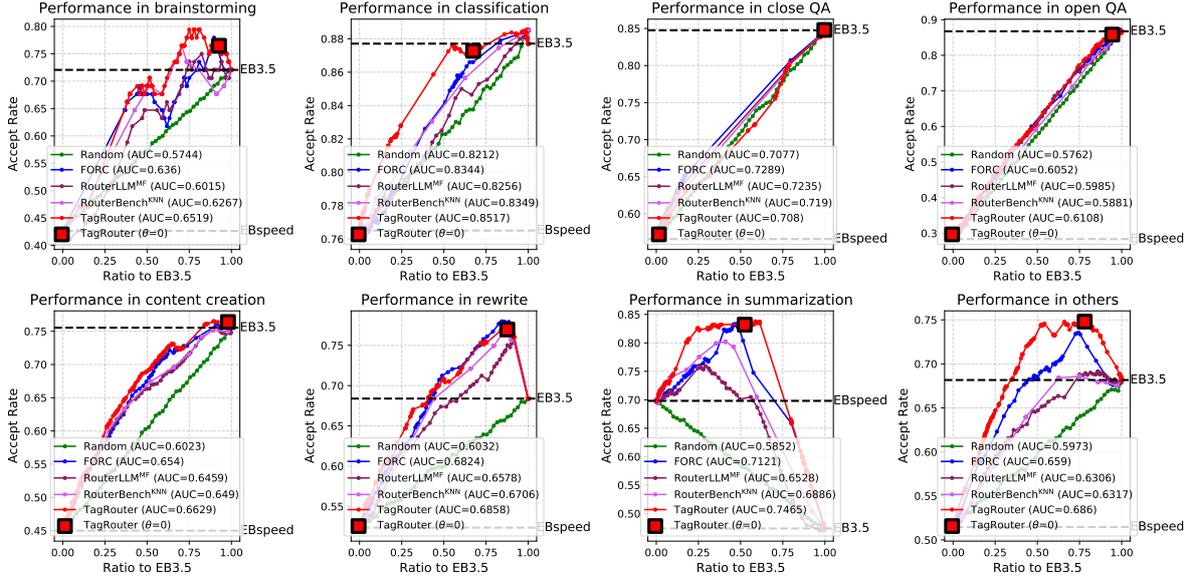

Figure 2: Comparison of TAGROUTER and the top three ranking existing routing methods across eight task categories in BCUQ dataset. The *ratio to EB3.5* represents the proportion of queries routed to EB3.5, where a higher ratio implies increased cost within the system. TAGROUTER outperforms baselines across most tasks.

**TAGROUTER outperforms baselines across most tasks.** Except for the close QA task, TAGROUTER achieves the highest AUC score in the remaining seven tasks. Notably, when its AR score exceeds that of EB3.5, TAGROUTER shows a clear advantage over baselines. Moreover, the threshold $\theta = 0$ selects a satisfactory value for the *ratio to EB3.5*, ensuring the system is cost-effective.

#### 5.2.3 Scaling TAGROUTER

The ability to ensemble additional LLMs into the routing system is critical to exploit the rapidly evolving model landscape effectively. Fig. 3 illustrates the performance of the TAGROUTER on the BCUQ dataset as the number of candidate models progressively increases from two to five.

**Expanding the model system leads to consistent performance improvements.** Specifically, as the number of candidate models increases from two to three and subsequently to five, the AUC score of the model system rises from 0.7610 to 0.7933, and further to 0.8043. This demonstrates that ensembling more models enhances the performance of the system. Moreover, the model system maintains a comparable AR score when operating with the threshold setting $\theta = 0$, while simultaneously reducing costs. These findings suggest that increasing the number of candidate models not only boosts performance but also improves cost-efficiency.

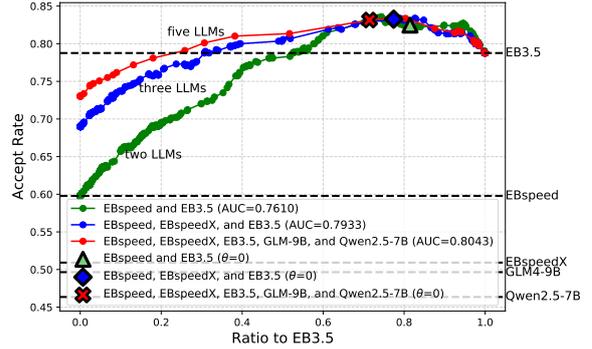

Figure 3: Scalability of TAGROUTER. Performance improves with more candidate models (from **two** to **three** to **five**), with enhanced AUC and cost-efficiency.

#### 5.2.4 Ablation Study

We conduct an ablation study on each component within every module of TAGROUTER to evaluate the performance of the routing system comprehensively. By systematically removing or modifying individual components, we analyze their respective contributions to the routing system.

**TAGGENERATOR.** *(i)* The proposed Hybrid Weight-Based Data Sampling algorithm (Alg. 2) enhances the performance of TAGGENERATOR. Experimental results (Tab. 8) show that a sampling ratio of 0.3 yields optimal performance. Moreover, the tag normalization component improves the performance of the routing system (Fig. 11). *(ii)* We evaluate Qwen2.5 and Llama3.2 series with varying parameter scales to balance performance and cost of the routing system (Tab. 9). Experimental results show that the Qwen2.5-0.5B is the best base model. *(iii)* We compare TAGGENERATOR against INSTAGGER, a model with 7 billion parameters to assess the complexity and diversity of the instruction data. Experimental results confirm the superior performance of TAGGENERATOR in the model routing field.

| Method | GLM4-9B and Qwen2-7B | | | EB3.5 and EBspeed | | | Average |
|---|---|---|---|---|---|---|---|
| | Alpaca | Dolly | BCUQ | Alpaca | Dolly | BCUQ | |
| RouteLLM<sup>MF</sup> (Ong et al., 2024) | 0.7142 | 0.7566 | 0.7626 | 0.6950 | 0.6475 | 0.7394 | 0.7192 |
| RouterBench<sup>KNN</sup> (Hu et al., 2024) | 0.7326 | 0.7583 | 0.7548 | 0.6978 | 0.6216 | 0.7515 | 0.7194 |
| FORC (Sakota et al., 2024) | 0.7384 | 0.7620 | 0.7659 | 0.7077 | 0.6700 | 0.7573 | 0.7336 |
| TAGROUTER | **0.7438** | **0.7623** | **0.7706** | **0.7239** | **0.7016** | **0.7610** | **0.7439** |

Table 3: Performance of TAGROUTER and top-3 baselines on Alpaca, Dolly and BCUQ datasets. Experiments are conducted separately with two candidate model groups: GLM-9B and Qwen2-7B, and EB3.5 and EBspeed. **Bold** numbers indicate the best results among all routing methods. TAGROUTER outperforms all baselines.

**TAGSCORER.** *(i)* The tag alignment component enhances the performance of the routing system (Fig. 11). *(ii)* In Ong et al. (2024), the values of $s_{win}$, $s_{tie}$, and $s_{loss}$ were set to 1, 1, and -1, respectively. However, we argue that the contribution of $s_{tie}$ to the candidate model should differ from that of $s_{win}$. Experimental results suggest that the optimal value for $s_{tie}$ is 0.15, as shown in Fig. 12.

**TAGDECIDER.** Fig. 13 illustrates the impact of different $\theta$ values on the model routing system. Experimental results show that the default setting of $\theta = 0$ yields satisfactory performance.

## 6 Discussions

**How does TAGROUTER perform among models with similar capabilities?** Fig. 6 presents the performance of TAGROUTER when GLM-9B and Qwen2.5-7B are selected as candidate models. Experimental results demonstrate that TAGROUTER effectively assigns different queries to GLM-9B and Qwen2.5-7B, validating its routing capability among models with high similarity.

**Can TAGGENERATOR generalize to other dataset?** Fig. 7 illustrates the performance of TAGROUTER, trained on the BCUQ dataset, when applied to the Alpaca and Dolly datasets. Results indicate that TAGROUTER identifies query characteristics effectively across diverse datasets. Moreover, it requires only a small number of labeled samples from the target dataset to further enhance its performance. Interestingly, even without dataset-specific optimization, TAGROUTER consistently outperforms existing routing methods that have been fine-tuned on the specific datasets, underscoring its strong generalization capability (Fig. 8).

**How should the threshold of TAGDECIDER be selected?** Extensive experiments indicate that the default setting of $\theta = 0$ is generally effective. For further optimization, Appx. G.2 presents a method for adapting $\theta$ to different datasets.

**How practical is TAGROUTER?** TAGROUTER is applicable to model routing across text generation tasks and benefits from a training-free manner. When new candidate models are added to the model system, only a small number of samples need to be annotated using the LLM-as-a-judge approach (Tab. 7 presents the performance under varying sample sizes). The capability features of new candidate models are then stored and quantified in a key-value format. This mechanism enables efficient expansion of the routing system without requiring retraining, ensuring adaptability to the rapidly evolving LLM ecosystem. Moreover, TAGROUTER consistently outperforms baseline methods across different datasets and candidate model groups (Tab. 3).

**How efficient is TAGROUTER?** In TAGROUTER, we utilize a 500MB TAGGENERATOR and a 33MB embedding model, with routing performed via simple key-value lookups. Compared to existing routing methods, this design offers a competitive advantage in computational efficiency and latency.

**Why does TAGROUTER exhibit superior performance?** As shown in Tab. 2 and Fig. 5, the four tag-based routing methods outperform 10 existing methods. We hypothesize that this superior performance stems from the ability of TAGGEN-

ERATOR to extract the core semantic features of potentially redundant, high-dimensional textual information and encode them into a structured set of tags (Appx. E.5 and Appx. E.6). This process can be seen as a form of automatic dimensionality reduction or feature abstraction, allowing routing models like TAGROUTER to focus on the main features. Therefore, the routing system achieves improved learning efficiency and performance.

## 7 Conclusions

In this work, we introduce TAGROUTER, a training-free routing method designed to scale with the growth of LLMs and handle open-domain text generation tasks. Extensive experimental evaluations demonstrate that TAGROUTER not only outperforms 13 baseline routing methods across a variety of datasets and tasks, but also exhibits strong adaptability and generalization. By dynamically orchestrating LLMs of varying scales and abilities, TAGROUTER allows users to benefit from high-performance LLM services without always relying on larger LLMs, reducing costs and improving efficiency of the system. Its practical design positions TAGROUTER as a promising solution for developing cost-efficient model systems.

## Limitations

**Language Capability.** The BCUQ dataset primarily comprises queries in Chinese and English, leading to the TAGGENERATOR that is limited to processing these two languages.

**Evaluation Methods.** *(i)* While the LLM-as-a-judge evaluation method may be less reliable than human evaluation, large-scale human evaluations are impractical due to the vast number of models, datasets, and experiments. Tab. 5 demonstrates a strong consistency between the two evaluation methods. *(ii)* Using a single model as the reference model $M_{\text{LLM}}$ may limit the advantages of crowd-sourcing approaches like Chatbot Arena. Evaluating the quality of LLM-generated responses using the Elo rating system to obtain more precise tag-score pairs could provide a more efficient solution and support scaling of the model system. We leave this avenue for future research.

## Ethical Statement

This work aims to provide a cost-efficient model routing method for inference in the era of LLMs. This method facilitates a more equitable distribution of LLM advancements, extending their benefits beyond well-resourced institutions to a wider range of users, promoting fairness and inclusivity within the NLP community. Furthermore, by dynamically selecting models rather than relying solely on larger LLMs, our method helps organizations reduce costs, lower carbon emissions, and support sustainable development.

# Appendix

## A Related Works

### A.1 Model Enhancement

Techniques such as fine-tuning (Chen et al., 2025a), Retrieval-Augmented Generation (RAG) (Lewis et al., 2020), and agentic LLMs (Qian et al., 2024) have been wisely used for improving model performance on specific tasks. However, these methods generally require additional training, domain-specific data, or intricate workflows (Chen et al., 2024b). In contrast, methods like Chain-of-Thought (CoT) (Wei et al., 2022), few-shot learning (Song et al., 2023), and prompt engineering (Ye et al., 2023) enhance performance without necessitating model training. Additionally, Mixture of Experts (MoE) approaches (Jacobs et al., 1991; Dai et al., 2024) enhance performance through intelligent routing, leveraging specialized expert modules within the model. Despite their utility, these methods do not fully exploit the synergistic potential of multiple models and model systems.

### A.2 LLM Tagging

Studies have demonstrated that capturing the semantic features of a task or query through tagging and supplying these tags to LLMs can effectively activate the various specialized capabilities of model. Tag-LLM (Shen et al., 2024) incorporates tags directly within the embedding layers as soft prompts, enhancing the specialized capabilities of model. Feldman et al. (2023) use tags to detect domain-external knowledge, reducing erroneous fabrications in LLMs. Further, Lu et al. (2024c) introduced INSTAGGER, an LLM with seven billion parameters tailored for generating tags in open domains, capable of assessing the diversity and complexity of instruction data to improve data sampling. ZOOTER (Lu et al., 2024b) employs INSTAGGER for adjusting biases in the off-the-shelf reward models to facilitate model routing. However, this method does not address the costs of using and retraining reward models. Unlike previous studies, this work introduces the lightweight TAGGENERATOR, specifically designed to facilitate model routing in a training-free manner.

## B Implementation Details

### B.1 Training TAGGENERATOR

We train TAGGENERATOR on the BCUQ dataset, sampled using Alg. 2, for one epoch to mitigate the risk of overfitting. We adopt Qwen2.5-0.5B as the base model and optimize it using the AdamW optimizer (Loshchilov and Hutter, 2019), with a maximum learning rate of $5e^{-5}$. A cosine learning rate schedule is employed, incorporating a 10% warm-up ratio. Training is performed on eight A100 80G GPUs, with a global batch size of 32. The maximum token length is set to 4096.

### B.2 Training Baselines

For baseline methods where an open-source model implementation is available, we directly use the off-the-shelf model. In cases where no such implementation is available, we replicate the model following the specifications provided in the original paper as similar as possible. For each baseline method, we perform multiple experimental configurations and report the best-performing results.

**FrugalGPT:** We extend the standard DistilBERT (Sanh et al., 2020) by adding a linear layer, which takes the final representation as input and produces a two-dimensional vector that encodes the correctness of the answer. The learning rate optimized via grid search is $1e^{-4}$.

**PairRanker:** We employ the off-the-shelf PairRanker from LLM-Blender (Jiang et al., 2023) to rank model-generated responses in pairs and route the query to the highest-ranked model. We conduct inference five times and report the result with the highest AUC score.

**Blending:** This method randomly selects a candidate model to respond to the query, enhancing response diversity. We conduct inference five times and report the result with the highest AUC score.

**RouteLLM$^{MF}$:** We first generate text embeddings using a pre-trained language model. Then, Singular Value Decomposition (SVD) (Klema and Laub, 1980) is applied to reduce the dimensionality of these embeddings. Finally, a logistic regression classifier is used for classification. We experiment with four embedding models: all-MiniLM-L12-v2 (Wang et al., 2021b), acge-text-embedding (Aspire, 2024), bge-base-en-v1.5, and bge-base-zh-v1.5 (Chen et al., 2024a), selecting all-MiniLM-L12-v2 as the best-performing model. The SVD dimensionality parameter is tuned via hyperparameter search, with the optimal dimension found to be 50.

**RouteLLM$^{SW}$:** We generate text embeddings using a pre-trained language model and classify them using a class-center similarity-based ranking method. After experimenting with four embedding models, we select all-MiniLM-L12-v2 as the best performer. The number of class centers optimized via grid search is 11.

**RouteLLM$^{\text{BERT}}$:** We employ BERT (Devlin et al., 2019) for text classification, incorporating an additional fully connected layer for binary classification. We use an entropy-based loss function for loss calculation and AdamW as the optimizer. The model is trained for two epochs. The learning rate optimized via grid search is $5e^{-5}$.

**RouteLLM$^{\text{LLM}}$:** We incorporate model identifiers as additional tokens in the vocabulary of Qwen2.5-0.5B, specifically adding `<Model_A>` and `<Model_B>`. LoRA (Hu et al., 2021) is applied to fine-tuning Qwen2.5-0.5B to enable LLM-based text classification. The optimal learning rate determined via grid search is $5e^{-4}$.

**RouteBench$^{\text{KNN}}$:** Text embeddings are generated using a pre-trained language model and classified using a KNN classifier. Among the four embedding models tested, acge-text-embedding performs best. The optimal $K$ value determined via grid search is 11.

**RouteBench$^{\text{MLP}}$:** We generate text embeddings using a pre-trained language model and classify them using an MLP. Among the four embedding models tested, bge-base-zh-v1.5 achieves the best performance. We experiment with different numbers of hidden layers (one, two, and three) and find that two hidden layers yield the best results.

**FORC:** We adopt transfer learning with DistilBERT, introducing two special tokens `<Model_A>` and `<Model_B>` in its vocabulary to differentiate between models and classification tasks. The optimal learning rate determined via grid search is $7e^{-5}$.

**RouteLLM$^{\text{MF}}$ w/TAGGENERATOR:** We use TAGGENERATOR as a feature extractor, select all-MiniLM-L12-v2 as the embedding model, and set the SVD dimensionality reduction parameter to 50.

**RouterBench$^{\text{KNN}}$ w/TAGGENERATOR:** We use TAGGENERATOR as a feature extractor, choose acge-text-embedding as the embedding model, and set the $K$ value to 11.

**FORC w/TAGGENERATOR:** We use TAGGENERATOR as a feature extractor. For DistilBERT, the learning rate is set to $7e^{-5}$.

## C Dataset

### C.1 BCUQ Details

The BCUQ dataset consists of 95,559 samples, categorized into eight distinct task categories. Fig. 4

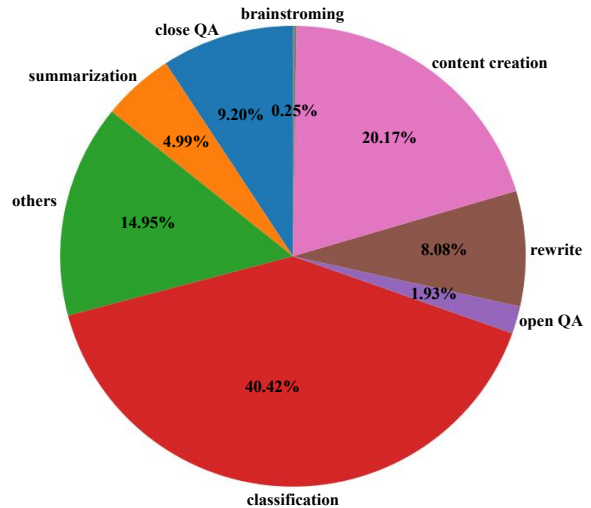

Figure 4: Task distribution in BCUQ.

shows the distribution of task types in BCUQ dataset. The classification of tasks is as follows:

**Brainstorming**: This task focuses on generating creative ideas or solutions to stimulate innovation.

**Classification**: This task involves the automatic categorization of text, including tasks like sentiment analysis and topic identification.

**Close QA**: This task requires the model to answer factual questions based on specific texts or knowledge bases.

**Open QA**: This task involves questions that do not have fixed answers, such as general knowledge questions or opinion-based queries.

**Content Creation**: In this task, the model is required to generate coherent and creative text, such as articles or advertising copy.

**Rewrite**: This task involves rephrasing or modifying a given text, such as transforming its style or optimizing its grammar.

**Summarization**: The goal of this task is to extract key information from long texts to produce concise summaries.

**Others**: This category encompasses tasks that do not belong to any of the previously defined categories, including but not limited to code generation and translation.

### C.2 Automatic and Human Evaluation on BCUQ

Given the cost and feasibility constraints associated with large-scale evaluations, this work employs the cost-efficient EB4.0 to assess the quality of responses generated by various models. To validate the reliability of the automated evaluation method, we randomly selected 50 samples from the BCUQ dataset and computed the Cohen's Kappa coefficient between EB4.0 and human evaluation results. The Cohen's Kappa coefficient measures the agreement between two evaluators, with values closer to 1 indicating higher consistency. Moreover, we evaluated the consistency between the GPT-4, human evaluation, and EB4.0 evaluation results.

| Dataset | Train Size | Validation Size | Test Size | Query Source |
|---------|-----------|-----------------|-----------|--------------|
| Alpaca  | 51,014    | -               | 988       | GPT-4        |
| Dolly   | 14,013    | -               | 998       | Databricks Employees |
| BCUQ    | 93,669    | 1,000           | 890       | LLM Service Usage |

Table 4: Dataset statistics for Alpaca, Dolly and BCUQ datasets. The training, validation, and test set sizes are reported alongside the sources of the query data.

Tab. 5 presents the Cohen's Kappa coefficient results between human and two LLMs. The results indicate that EB4.0 exhibits a high level of consistency with both human evaluation and the GPT-4. Thus, the use of EB4.0 for automated evaluation is considered reliable.

| Comparison | Cohen's Kappa Value |
|------------|---------------------|
| Human vs. EB4.0 | 0.79 |
| Human vs. GPT-4 | 0.75 |
| EB4.0 vs. GPT-4 | 0.71 |

Table 5: Cohen's Kappa results between human and two LLMs. EB4.0 exhibits a high Cohen's Kappa value. One of the authors served as the human annotator.

### C.3 Dataset Statistics

This study utilizes the Alpaca (Wang et al., 2023), Dolly (Conover et al., 2023), and BCUQ datasets. The hyperparameters for TAGROUTER were optimized based on experiments conducted on the BCUQ dataset. The BCUQ dataset, sourced from LLM service usage, is more representative of open-domain text generation tasks compared to Alpaca and Dolly, thereby offering a closer reflection of real-world user demands and expectations for LLM capabilities. A detailed statistical summary of these datasets is provided in Tab. 6.

## D Additional Experiments in TAGROUTER

### D.1 Performance Comparison of TAGROUTER and Baselines

Fig. 5 presents supplementary results that complement Tab. 2, showing the performance of TAGROUTER and baseline methods on the BCUQ dataset as they vary with the *ratio to EB3.5*. The results demonstrate that TAGROUTER consistently outperforms all baseline methods in terms of AUC. Notably, in the high-gain region where the AR value surpasses that of EB3.5, TAGROUTER exhibits an even more significant performance advantage. This observation underscores the effectiveness of TAGROUTER in enhancing system performance through ensembling multiple models.

### D.2 Routing Capability Among Comparable LLMs

Significant differences in parameter sizes often lead to performance disparities. This has made model routing based on parameter size a widely studied topic (Aggarwal et al., 2023; Chen et al., 2023; Yue et al., 2024; Lee et al., 2024). However, we argue that even among models with similar parameter sizes, variations in training data, model architectures, and training methods can still lead to notable performance differences. In some cases, these variations may result in complementary strengths across specific tasks. Therefore, investigating efficient routing methods for LLMs with comparable parameter sizes is important.

To evaluate the routing capability of TAGROUTER in such scenarios, we selected GLM4-9B and Qwen2.5-7B as candidate models. These models not only have comparable parameter sizes but also exhibit similar performance on both Chinese and English comprehension tasks, as assessed by the CMMLU (Li et al., 2023) and MMLU (Hendrycks et al., 2021) benchmarks. We further evaluated their performance on the BCUQ dataset, with the experimental results presented in Fig 6, where the blue curve represents the performance of TAGROUTER under this setting.

The results demonstrate that TAGROUTER effectively improves the AR score of the model system while simultaneously reducing computational cost, thereby enhancing system efficiency. This further validates the effectiveness and applicability

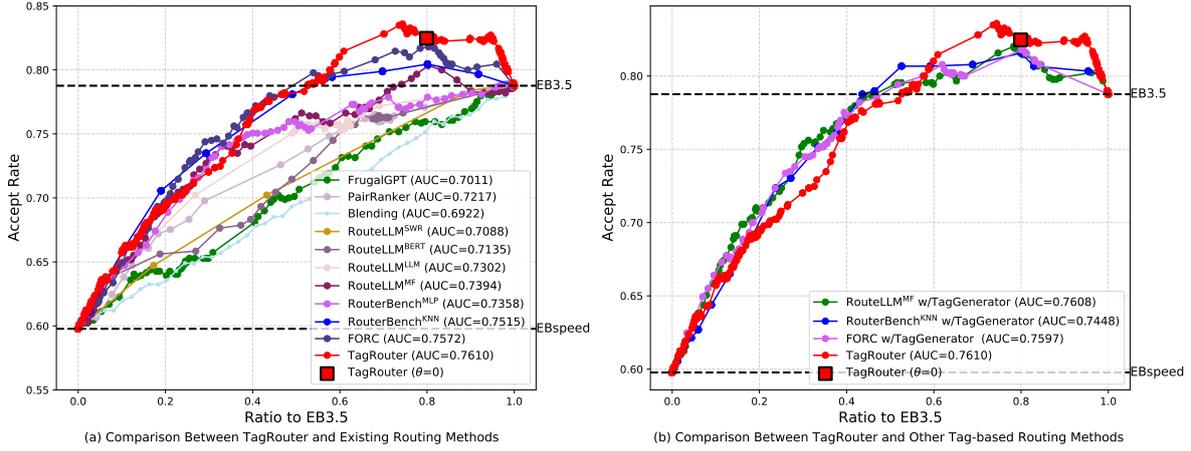

Figure 5: Performance comparison of TAGROUTER and the baseline methods on BCUQ dataset. TAGROUTER outperforms all baselines. (a) Comparison between TAGROUTER and the top three existing routing methods. (b) Comparison between TAGROUTER and other tag-based routing methods introduced in Sec. 5.1.

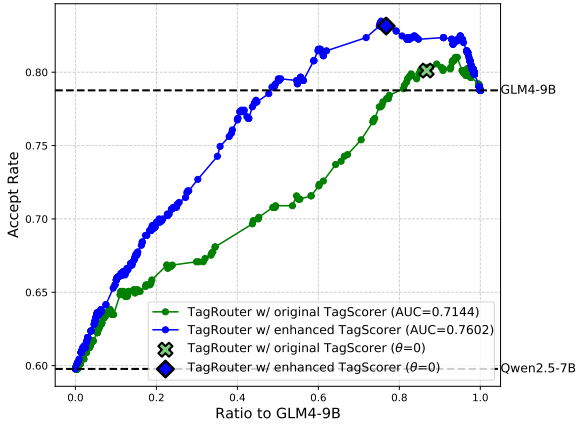

Figure 6: Performance of TAGROUTER on BCUQ dataset. The candidate LLMs are GLM4-9B and Qwen2.5-7B. "w/ original TAGSCORER" denotes the use of tag-score pairs generated by EB3.5 and EBspeed as capability representations, while "w/ enhanced TAGSCORER" refers to the use of tag-score pairs generated by GLM4-9B and Qwen2.5-7B.

of TAGROUTER in routing LLMs with comparable capabilities.

### D.3 Generalization to Unseen LLMs

Routing without the need for labeled samples from unseen LLMs is critical for the practical applicability of routing systems. In this work, we selected GLM4-9B and Qwen2.5-7B as candidate models. To assess generalization performance, we use tag-score pairs generated by EB3.5 as the capability representation for GLM4-9B and tag-score pairs generated by EBspeed for Qwen2.5-7B. These representations were then used to evaluate the ability of TAGROUTER to generalize on the BCUQ dataset.

The experimental results are presented in Fig. 6, where the green curve corresponds to the scenario without labeled samples.

The results indicate that although the AUC scorer in the no-labeled-sample setting (green curve) is lower than in the labeled-sample setting (blue curve), TAGROUTER still significantly enhances the performance of the model system. This suggests that TAGROUTER has implicitly learned to differentiate between complex and simple queries during training, enabling it to dynamically select the appropriate LLM for inference based on task complexity.

### D.4 Generalization to Other Benchmarks

To evaluate the generalization capability of TAGROUTER, we trained the model on the BCUQ dataset and assessed its performance on the Alpaca (Wang et al., 2023) and Dolly (Conover et al., 2023) datasets. The experimental results are illustrated in Fig. 7, where the green curve depicts the accept rate as a function of the ratio to EB3.5. The results demonstrate that TAGROUTER effectively enhances system performance compared to using an individual model, further validating its generalization ability across diverse datasets.

To further optimize system performance, we aggregate the tag-scores of candidate models computed on Alpaca and Dolly datasets with those obtained from BCUQ. By incorporating this enhancement strategy, the experimental results represented by the blue curve in Fig. 7, exhibit a significant improvement in model system performance. This finding reinforces the scalability and adaptability

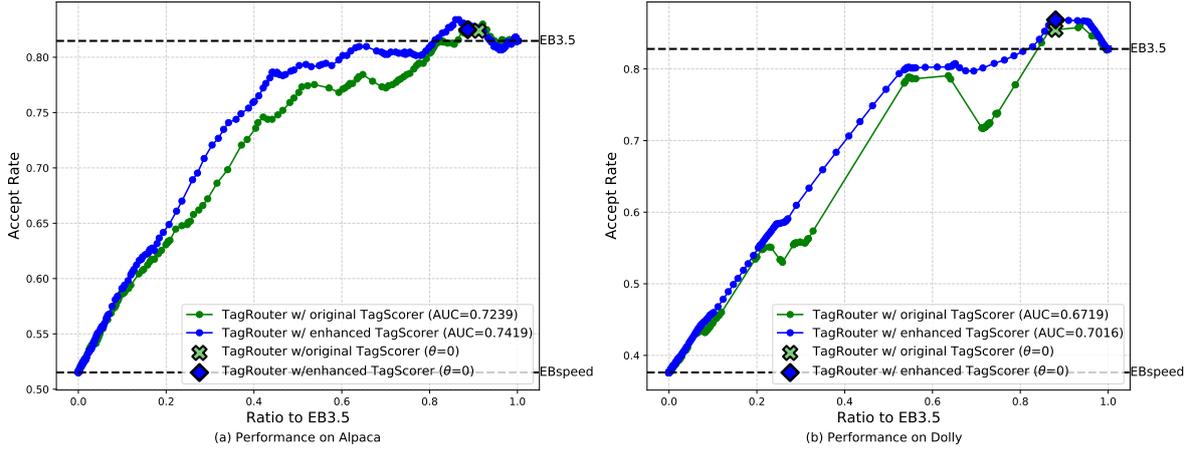

Figure 7: Performance of TAGROUTER on Alpaca and Dolly datasets. Candidate models include EB3.5 and EBspeed. "w/ original TAGSCORER" refers to routing based solely on tag-scores computed from the BCUQ dataset, whereas "w/ enhanced TAGSCORER" incorporates tag-scores computed from the training sets of the target evaluation datasets (Alpaca and Dolly) in addition to those from BCUQ.

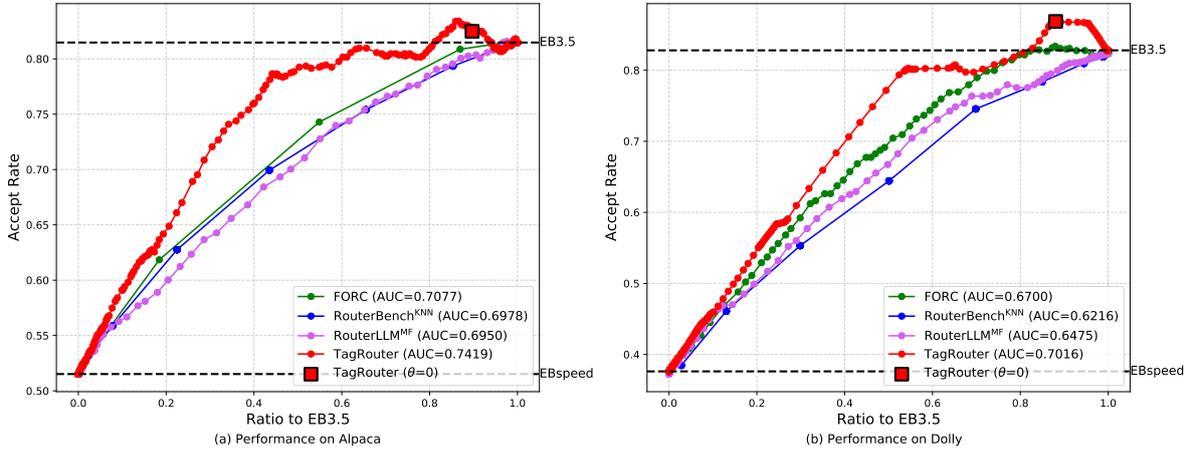

Figure 8: Performance comparison of TAGROUTER and the top three ranking existing routing methods on Alpaca and Dolly datasets. TAGROUTER outperforms all baselines.

of TAGROUTER as a training-free routing method.

Furthermore, Fig. 8 presents a comparative analysis between TAGROUTER and the top three ranking existing routing methods on the Alpaca and Dolly datasets. The results indicate that TAGROUTER is the only method capable of achieving a AR score that surpasses all individual candidate models, further substantiating its effectiveness in model routing tasks.

### D.5 Analysis of TAGROUTER Across Different Benchmarks

By examining Fig. 5 and Fig. 8, we observe notable variations in the effectiveness of model routing methods across different datasets in terms of improving routing system performance and surpassing all individual candidate models. For instance, on the BCUQ dataset, both TAGROUTER and baseline methods significantly enhance model system performance. However, achieving comparable performance improvements on the Alpaca and Dolly datasets proves to be more challenging. Analyzing this phenomenon provides deeper insights into the applicability of routing methods in diverse real-world scenarios.

Tab. 6 presents the key statistics of the three datasets alongside the performance of TAGROUTER. The following observations can be drawn: longer queries tend to contain a greater number of tags, which serve as representations of user intent. For example, Alpaca exhibits the fewest tags and the lowest PAUC score, whereas BCUQ contains the highest number of tags, corresponding to the highest PAUC score. This suggests that a greater number of tags facilitates a more distinctive query representation, enabling the routing system to more effectively allocate queries to the most appropriate model.

### D.6 Impact of Training Data Size

Tab. 7 presents the performance of TAGROUTER on the BCUQ dataset when trained with varying amounts of data. The experimental results indicate that even with only 100 training samples, the AR score of the model system improves by 0.86%. As the training data size increases, system performance continues to improve, suggesting that a larger training samples further enhances the effectiveness of the routing system.

## E TAGGENERATOR

### E.1 Algorithms for Developing TAGGENERATOR

---

**Algorithm 1** Iterative Reduction of Tags within Clusters for a Set of Queries

---

**Input** : A set of queries $Q = \{q_1, q_2, \ldots, q_n\}$, each associated with a set of tags $\mathcal{T}(q)$
**Output :** The reduced set of tags after clustering and reduction for all queries

EncodeTags($\mathcal{T}(q)$) **return** *Normalized embeddings of $\mathcal{T}(q)$*
DBSCANCluster($E_q$) **return** *Clusters based on the distance matrix derived from $\mathcal{E}_q$*
ReduceTags($C$) **while** $|C| > 2$ **do**
    Remove the tag with the least cumulative similarity within $C$
**end**
**return** $C$
ReducedTags $\leftarrow \emptyset$
**foreach** *query $q \in Q$* **do**
    $E_q \leftarrow$ EncodeTags($\mathcal{T}(q)$)
    $Clusters \leftarrow$ DBSCANCluster($E_q$)
    **foreach** *cluster $C$ in Clusters* **do**
        $ReducedC \leftarrow$ ReduceTags($C$)
        ReducedTags $\leftarrow$ ReducedTags $\cup$ $ReducedC$
    **end**
**end**
**return** ReducedTags

---

**Algorithm 2** Hybrid Weight-Based Data Sampling

---

**Input** : Training dataset $\mathcal{D}$ with associated tags, sampling ratio $\alpha \in (0, 1]$
**Output :** Sampled training dataset $\mathcal{D}_{\text{sampled}}$

**Step 1: Compute Hybrid Weights for Tags** Compute frequency $f_t$ of each tag $t \in \mathcal{D}$
**foreach** *tag $t \in \mathcal{D}$* **do**
    Compute hybrid weight:
$$w_t^{\text{hybrid}} \leftarrow \frac{1}{f_t} + \log\left(\max_{t \in \mathcal{T}} f_t\right) - \log(f_t)$$

    Assign $w_t^{\text{hybrid}}$ to corresponding entries in $\mathcal{D}$
**end**

**Step 2: Normalize Weights** Compute total weight:
$$\mathcal{W}_{\text{total}} \leftarrow \sum_{t \in \mathcal{D}} w_t^{\text{hybrid}}$$

**foreach** *data point $d \in \mathcal{D}$* **do**
    Normalize weight:
$$w_d^{\text{normalized}} \leftarrow \frac{w_d^{\text{hybrid}}}{\mathcal{W}_{\text{total}}}$$

**end**

**Step 3: Perform Weighted Sampling**
Determine the number of samples to draw based on $\alpha$:
$$n = \lceil \alpha \cdot |\mathcal{D}| \rceil$$

Initialize $\mathcal{D}_{\text{sampled}} = \emptyset$ with capacity $n$ (sampled dataset)
**for** $i = 1$ **to** $n$ **do**
    Sample a data point $d_i$ from $\mathcal{D}$ with probability proportional to $w_{d_i}^{\text{normalized}}$
    Append $d_i$ to $\mathcal{D}_{\text{sampled}}$
    Remove $d_i$ from $\mathcal{D}$ to avoid re-sampling
**end**
**return** $\mathcal{D}_{\text{sampled}}$

---

### E.2 Grid Search for the Best $\alpha$

We adopted the knowledge distillation method to transfer knowledge from the large model to the small model for training the TAGGENERATOR. Specifically, we first used EB4.0 to generate tags corresponding to queries, which were then used to construct the instruction dataset $\mathcal{D}$ to fine-tune base model (Qwen2.5-0.5B in this experiment). However, we observed a significant class imbalance in the instruction dataset. Therefore, we applied Alg. 2 for sampling, where the sampling ratio $\alpha$ determines the number of training samples. To find the optimal $\alpha$, we performed grid search for hyperparameter tuning. Tab. 8 shows the consistency and diversity evaluation results between the tags generated by TAGGENERATOR and those generated by EB4.0 for different values of $\alpha$.

| Dataset | Average Query Tokens | Average Tag Count | PAUC |
|---|---|---|---|
| Alpaca | 18.67 | 1.95 | 0.14 |
| Dolly | 107.13 | 2.19 | 0.50 |
| BCUQ | 329.98 | 3.33 | 1.46 |

Table 6: Basic statistics of the three datasets and the performance of TAGROUTER. Query token counts are computed using the EBspeed tokenizer, tag numbers are generated by TAGGENERATOR, and PAUC score represents the performance of TAGROUTER on the respective dataset.

| Category | Method | Performance at Max AR | | | | AUC(%)↑ | PAUC(%)↑ |
|---|---|---|---|---|---|---|---|
| | | AR(%)↑ | Uplift(%)↑ | Cost↓ | Rank↓ | | |
| Single LLM | EBspeed | 59.78 | -24.1 | 2.01 | 1.212 | - | 0 |
| | EB3.5 | 78.76 | 0 | 13.49 | 1.400 | - | 0 |
| Training Data | 100 | 79.44 | 0.86 | 12.49 | 1.206 | 71.36 | 0.01 |
| | 300 | 80.79 | 2.58 | 13.02 | 1.192 | 73.18 | 0.19 |
| | 500 | 80.90 | 2.72 | 12.82 | 1.191 | 73.45 | 0.22 |
| | 1,000 | 81.01 | 2.86 | 12.78 | 1.190 | 73.95 | 0.27 |
| | 3,000 | 81.24 | 3.15 | 12.55 | 1.188 | 75.28 | 0.75 |
| | 5,000 | 82.25 | 4.43 | 12.56 | 1.178 | 75.46 | 0.97 |
| | 10,000 | 82.58 | 4.85 | 12.56 | 1.174 | 75.48 | 0.95 |
| | 30,000 | 83.37 | 5.85 | 11.35 | 1.166 | 75.95 | 1.29 |
| | 50,000 | 83.26 | 5.71 | 11.32 | 1.167 | 75.90 | 1.30 |
| | 70,000 | 83.48 | 5.99 | 11.32 | 1.165 | 76.00 | 1.40 |
| | 93,669 | 83.60 | 6.15 | 11.17 | 1.164 | 76.10 | 1.46 |

Table 7: Performance of TAGROUTER on BCUQ dataset with different size of training data.

**Consistency.** From the F1-score results, we can see that as $\alpha$ increases, the consistency between the tags generated by TAGGENERATOR and those generated by EB4.0 consistently improves. This phenomenon indicates that, as the training data increases, TAGGENERATOR better learns the pattern of tags generated by EB4.0, leading to a higher match rate in the generated tags.

**Diversity.** From the inter-rate results, we observe a trend where the diversity of the generated tags first increases and then decreases as $\alpha$ increases. When $\alpha$ is small, a moderate increase in $\alpha$ enhances the model ability to learn the tag generation pattern, thus improving the diversity of the generated tags. However, as $\alpha$ grows further, the proportion of high-frequency tags in the training data increases, leading to overfitting on these high-frequency tags, which in turn reduces the diversity of the generated tags.

When $\alpha = 0.10$, TAGGENERATOR achieves both high consistency and diversity. Therefore, we select $\alpha = 0.10$ as the final parameter for training the TAGGENERATOR.

### E.3 Selecting Base Model

We selected the base models suitable for TAGGENERATOR from the Qwen2.5 and Llama3.2 series. Tab. 9 shows the performance of the TAGGENERATOR trained with different base models at $\alpha = 10$ in terms of consistency, diversity, and routing performance. Here, routing performance refers to the AUC score of TAGROUTER on the BCUQ dataset when EB3.5 and EBspeed are used as candidate models.

As the model parameter size increases, consistency, diversity, and routing performance all show an upward trend. When using Qwen2.5-0.5B as the base model, the routing system not only performs excellently but also maintains low cost and latency due to its small parameter size. Therefore, Qwen2.5-0.5B is chosen as the final base model.

| $\alpha$ | Accuracy | Precision | Recall | F1-Score | Inter Rate |
|---|---|---|---|---|---|
| 0.03 | 31.84 | 46.74 | 49.96 | 48.30 | 0.6340 |
| 0.05 | 37.82 | 53.46 | 56.40 | 54.89 | 0.7448 |
| 0.08 | 32.67 | 48.45 | 50.08 | 49.25 | 0.8144 |
| 0.10 | 40.60 | 55.85 | 59.78 | 57.75 | 0.8686 |
| 0.20 | 41.08 | 56.61 | 59.97 | 58.24 | 0.8325 |
| 0.30 | 40.57 | 56.48 | 59.03 | 57.73 | 0.7887 |
| 0.40 | 43.71 | 59.39 | 62.34 | 60.83 | 0.7809 |
| 0.50 | 48.72 | 65.24 | 65.80 | 65.52 | 0.5600 |
| 0.80 | 45.55 | 61.73 | 63.47 | 62.59 | 0.4716 |

Table 8: Consistency and diversity evaluation results between tags generated by TAGGENERATOR and EB4.0. Accuracy, Precision, recall, and F1-score reflect the consistency between the tags generated by TAGGENERATOR and those generated by EB4.0. The inter rate metric measures the proportion of tag types generated by TAGGENERATOR in the EB4.0 tag set, which is used to evaluate the diversity of the generated tags. $\alpha = 0.10$ is the best.

| Base Model | Accuarcy | Precision | Recall | F1-Score | Inter Rate | AUC |
|---|---|---|---|---|---|---|
| Qwen2.5-0.5B | 40.60 | 55.85 | 59.78 | 57.75 | 0.8686 | 76.10 |
| Qwen2.5-1.5B | 40.77 | 55.78 | 60.23 | 57.92 | 0.9072 | 77.14 |
| Qwen2.5-3B | 40.11 | 55.12 | 59.56 | 57.25 | 0.8943 | 76.24 |
| Qwen2.5-7B | 41.00 | 55.79 | 60.72 | 58.15 | 0.8918 | 77.48 |
| Llama3.2-1B | 39.83 | 54.99 | 59.10 | 56.97 | 0.8660 | 76.03 |
| Llama3.2-3B | 40.87 | 55.69 | 60.57 | 58.03 | 0.8969 | 77.26 |

Table 9: Consistency, diversity, and routing performance of TAGGENERATOR trained with different base models.

### E.4 Compare TAGGENERATOR with INSTAGGER

INSTAGGER is an LLM with seven billion parameters designed for generating open-domain tags, and it can quantify the diversity and complexity of instruction data. This work compares the performance of the TAGROUTER using INSTAGGER for tag generation with the standard TAGROUTER (using TAGGENERATOR). Tab. 10 shows that TAGGENERATOR outperforms INSTAGGER across all metrics and can improve the performance of the model system more effectively.

### E.5 Win/Tie/Loss Distribution for Tags

Analyzing the contribution of various tags to the final model selection in the routing system helps us understand the role tags play in model routing decisions. Tags were selected from the tag set based on the proportion of the sum of "win" and "tie" counts relative to the total count of "win," "tie," and "loss" for each tag in the pairwise comparison results. We present the top 10 and bottom 10 tags, along with the distribution of pairwise comparison results, shown in Fig. 9 and Fig. 10. The pairwise comparison results were obtained using the LLM-as-a-judge method on the BCUQ dataset, which evaluates the quality of responses generated by EBspeed and EB3.5.

**Tags play a important role in model routing.** For queries associated with tags in Fig. 9, the routing system should select EBspeed as the final model. For example, when the tag "Medical Report" is generated, selecting EBspeed results in an AR score (sum of "win" and "tie") of 100%. Conversely, for queries corresponding to tags in Fig. 10, the system should select EB3.5. This "fine-grained classification" based on tags is challenging to achieve with predefined task categories.

**Tags with similar semantics contribute similarly to model routing.** In Fig. 9, we observe that tags related to experience (e.g., "Product Sales Experience," "Product Identification," "Experience Analysis") exhibit consistent contributions to the performance of EBspeed and EB3.5 on queries containing experience-related semantic features. Specifically, EBspeed performs better on these queries. Similarly, in Fig. 10, EB3.5 is more effective at handling queries related to travel, indicating that tags are interpretable in terms of model capabilities.

| Category | Method | Performance at Max AR | | | | AUC(%)↑ | PAUC(%)↑ |
|---|---|---|---|---|---|---|---|
| | | AR(%)↑ | Uplift(%)↑ | Cost↓ | Rank↓ | | |
| Single LLM | EBspeed | 59.78 | -24.1 | 2.01 | 1.212 | - | 0 |
| | EB3.5 | 78.76 | 0 | 13.49 | 1.400 | - | 0 |
| TAGROUTER | INSTAGGER | 82.47 | 4.71 | 12.31 | 1.175 | 74.18 | 1.13 |
| | TAGGENERATOR | 83.60 | 6.15 | 11.17 | 1.164 | 76.10 | 1.46 |

Table 10: Performance comparison between TAGROUTER using INSTAGGER (7B) and TAGGENERATOR (0.5B). TAGGENERATOR outperforms INSTAGGER across all metrics.

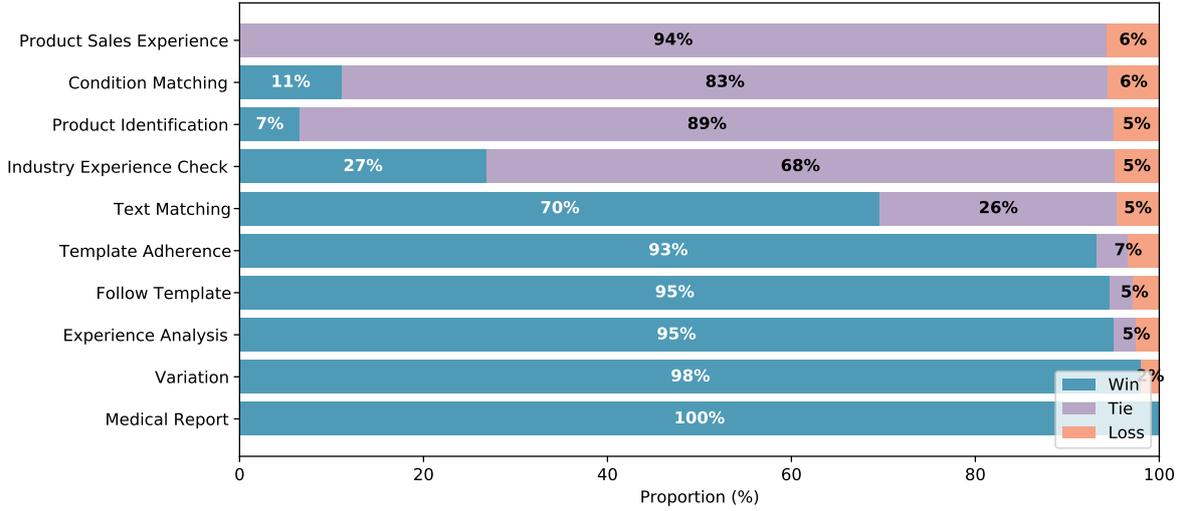

Figure 9: Win/Tie/Loss distribution for the top 10 tags.

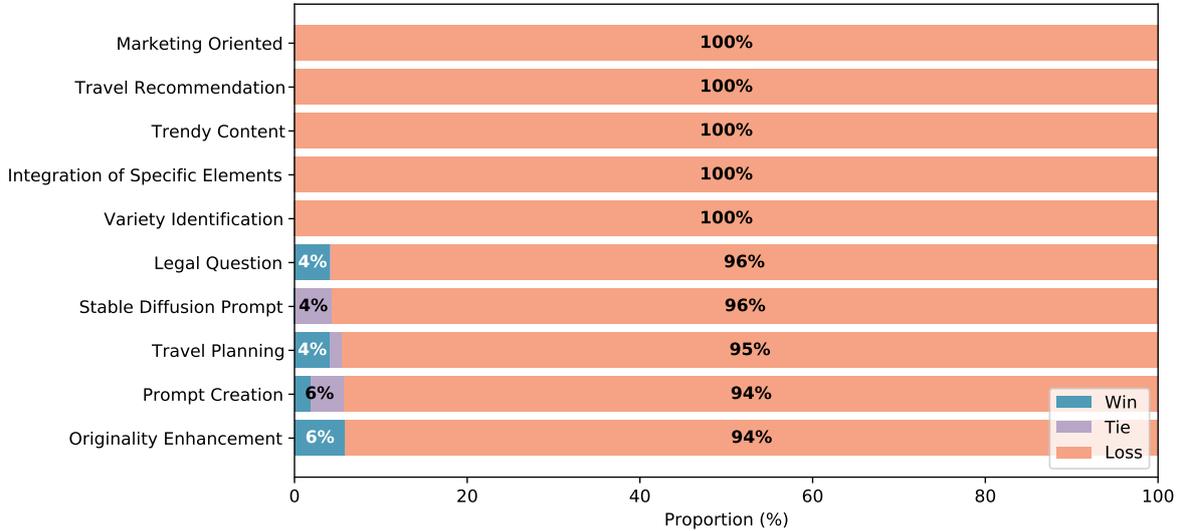

Figure 10: Win/Tie/Loss distribution for the bottom 10 tags.

### E.6 Cases of TAGGENERATOR

Tab. 11, 12, and 13 present the tagged cases from the Alpaca, Dolly, and BCUQ datasets, with tags generated by TAGGENERATOR. The tags accurately reflect user intentions.

## F TAGSCORER

### F.1 Impact of Tag Normalization and Alignment

To enhance the performance of TAGSCORER, we adopt the tag set obtained through tag normaliza-tion, followed by an embedding-based tag alignment procedure. These methods strengthens the robustness and generalization ability of the routing system. As illustrated in Fig. 11, both tag normalization and tag alignment enhances the performance of model system. Furthermore, we observe that after applying tag normalization, the effect of tag alignment on the AUC score is minimal, with only a marginal increase of 0.0001. This finding suggests that when optimizing for low-latency responses, tag alignment can be omitted while maintaining the AR score of the model system within a satisfactory range.

| Query | Tag |
| --- | --- |
| Describe a process of making crepes. | Text Generation, Process Description |
| Given the parameters of a triangle, find out its perimeter. Side 1 = 4, Side 2 = 6, Side 3 = 8 | Geometry, Problem Solving |
| Rewrite the sentence so that it's in the present tense: She had worked at the company for the past 3 years. | Text Rewriting, Language Style |

Table 11: Cases from Alpaca dataset tagged by TAGGENERATOR.

| Query | Tag |
| --- | --- |
| Identify which instrument is string or woodwind: Panduri, Zurna. | Text Classification, Knowledge Application |
| Who is Thomas Jefferson? Please answer the above question based on the following context: Thomas Jefferson (April 13, 1743 – July 4, 1826) was an American statesman, diplomat, lawyer, architect, philosopher, and Founding Father who served as the third president of the United States from 1801 to 1809. Among the Committee of Five charged by the Second Continental Congress with authoring the Declaration of Independence, Jefferson was the Declaration's primary author. Following the American Revolutionary War and prior to becoming the nation's third president in 1801, Jefferson was the first United States secretary of state under George Washington and then the nation's second vice president under John Adams. | Question Answering, Fact based Response |
| You are a master of marketing copy, tasked with creating a catchy slogan for a product named "One-Stop Website Solutions." The product's strengths are professionalism, ease, cost-effectiveness, and superior post-sales service. Try to emphasize these keywords: "professional team, dedicated post-sales support." | Text Generation, Advertising, Markdown Formatting, Keyword Incorporation |

Table 12: Cases from Dolly dataset tagged by TAGGENERATOR.

### F.2 Grid Search for the Best $s_{\text{tie}}$

In Ong et al. (2024), the values of $s_{\text{win}}$, $s_{\text{tie}}$, and $s_{\text{loss}}$ are set to 1, 1, and -1, respectively. We hypothesize that when the generated response results in a "tie" during pairwise comparisons with a value of $s_{\text{tie}}$, it should not be treated the same as $s_{\text{win}}$. Instead, it should lie within a range between 0 and 1. The experimental results, as shown in Fig. 12, suggest that the model achieves optimal performance when $s_{\text{tie}}$ is set to 0.15.

## G Additional Experiments in TAGDECIDER

### G.1 Performance at Different Values of $\theta$

The design of TAGDECIDER aims to enable routing system achieve the highest AR score when $\theta = 0$. Fig. 13 shows the performance of the model system across various values of $\theta$. Experiments conducted on three datasets demonstrate that the default setting of $\theta = 0$ is an satisfactory choice. In this configuration, the model system not only outperforms any individual model in AR score, but also incurs lower costs compared to the method of routing all queries to EB3.5.

| Query | Tag |
|---|---|
| Translate the following text into Chinese: To cool down, a snake moves into the shade. | Translation |
| Your task: Extract the core keywords from the input content and output them in the required format.<br>Requirements:<br>1. The extracted keywords should represent the core intent of the sentence.<br>2. The output should strictly follow the required format without any unrelated text.<br>3. Only output the required JSON format, without using markdown formatting.<br>Input content: What should I do if I catch a cold? | Keyword Extraction, Output Formatting, Text Processing |
| You are a master of marketing copy, tasked with creating a catchy slogan for a product named "One-Stop Website Solutions." The product's strengths are professionalism, ease, cost-effectiveness, and superior post-sales service. Try to emphasize these keywords: "professional team, dedicated post-sales support." | Text Generation, Advertising, Markdown Formatting, Keyword Incorporation |

Table 13: Cases from BCUQ dataset tagged by TAGGENERATOR.

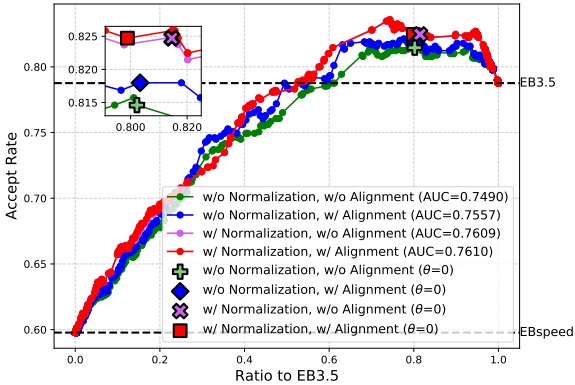

Figure 11: Impact of tag normalization and tag alignment on the performance of the routing system.

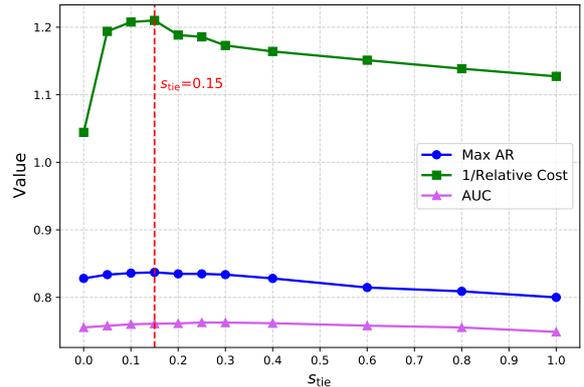

Figure 12: Impact of different $s_{\text{tie}}$ values on the performance of the model system. "1/Relative Cost" refers to the inverse of the normalized cost when the AR reaches its maximum value.

As $\theta$ decreases, the routing system increasingly prioritizes cost and routes more queries to the more affordable EBspeed, thereby reducing the system cost. However, when $\theta > 0$, further increasing $\theta$ results in some queries that should have been routed to EBspeed being incorrectly assigned to EB3.5, causing a degradation in performance. Thus, by dynamically adjusting $\theta$, we can achieve an optimal trade-off between performance and cost.

### G.2 Method for Best $\theta$ Selection

As shown in Fig. 13, while the default setting $\theta = 0$ is effective, it is not always the best value in different datasets. To identify the best $\theta$ tailored to the specific characteristics of different datasets, we employed a grid search method to evaluate the performance model system on the training set for various values of $\theta$, selecting the best value $\theta^*$. Specifically, we randomly sampled 1000 instances from the training sets of the Alpaca, Dolly, and BCUQ datasets to determine the best $\theta^*$. The results of this search algorithm are presented in Tab. 14. Experimental results show that the proposed method significantly improves the selection of the optimal $\theta$ across all three datasets. This method allows for dynamic selection of $\theta^*$ based on the unique characteristics of different datasets.

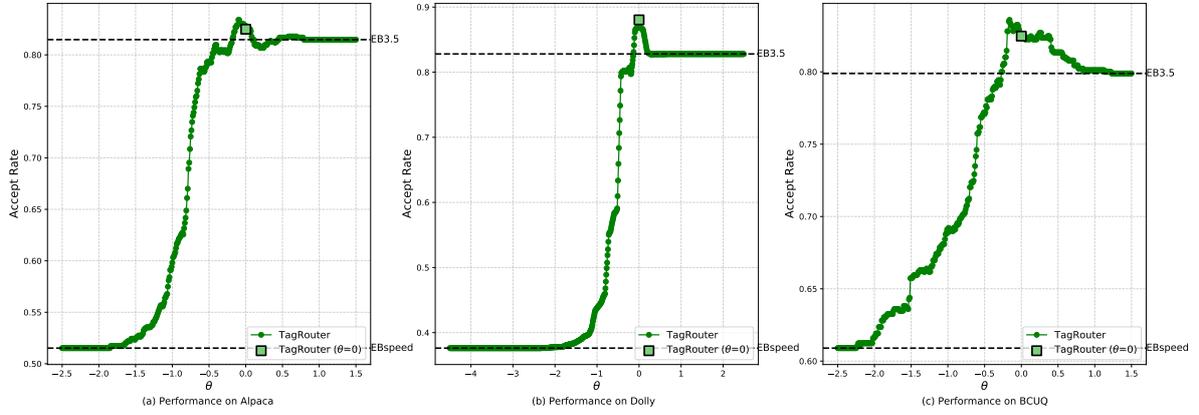

Figure 13: Performance of TAGROUTER on the Alpaca, Dolly, and BCUQ datasets for various values of $\theta$.

| Dataset | $\theta$ | Performance at Max AR | | | |
|---|---|---|---|---|---|
| | | AR(%)↑ | Uplift(%)↑ | Cost↓ | Rank↓ |
| Alpaca | $\theta=0$ | 82.49 | 1.25 | 12.81 | 1.175 |
| | $\theta=\theta^*$ | 86.64 | 6.35 | 12.63 | 1.166 |
| Dolly | $\theta=0$ | 86.67 | 4.68 | 15.34 | 1.131 |
| | $\theta=\theta^*$ | 86.86 | 4.91 | 14.35 | 1.132 |
| BCUQ | $\theta=0$ | 82.47 | 4.71 | 11.67 | 1.175 |
| | $\theta=\theta^*$ | 83.60 | 6.15 | 11.17 | 1.164 |

Table 14: Performance of TAGROUTER on the Alpaca, Dolly, and BCUQ datasets for $\theta=0$ and $\theta=\theta^*$. $\theta=\theta^*$ is more cost-effient than $\theta=0$.

## H  Prompt Template

**Prompt for Tag Generation Using EB4.0**

```
[System]
You are an instruction tagging system designed to provide coarse
    -grained tags for human instructions, aiming to identify and
    analyze the semantics and intentions of instructions through
    these tags.

Please provide coarse-grained tags, such as "Text Generation", "
    Spelling and Grammar Check", and "Cosplay", to clearly
    describe the main intentions of the provided instruction.
    These tags will aid in the quantitative analysis of the
    diversity and complexity of instructions. Here is an
    instruction:
[begin]
```json
{{
    \"instruction\": \"{prompt}\",
}}
```
[end]

Your response should include a list of tag titles and a brief
    explanation for each tag. Your response must strictly follow
    the JSON format below. Please respond in English.
[Output Format]
```json
[
    {{
        \"tag\": str,
        \"explanation\": str
    }}
]
```
```

**Prompt for Tag Generation Using TAGGENERATOR**

```
[System]
You are an expert text tag extractor. Your task is to identify
    tags that readers should focus on while engaging with the
    user query below.

[User Query]
{Input}
```

**Prompt for Pairwise Comparison of Model Responses Using EB4.0/GPT-4**

[System]
Please act as an impartial judge and evaluate the quality of responses provided by two AI assistants to the user question displayed below. Your evaluation should consider correctness and helpfulness. You will be given assistant A's answer, and assistant B's answer. Your job is to evaluate which assistant's answer is better. You should independently solve the user question step-by-step first. Then compare both assistants' answers with your answer. Identify and correct any mistakes. Avoid any position biases and ensure that the order in which the responses were presented does not influence your decision. Do not allow the length of the responses to influence your evaluation. Do not favor certain names of the assistants. Be as objective as possible. After providing your explanation, output your final verdict by strictly following this format: "A" if assistant A is better, "B" if assistant B is better, and "C" for a tie. Please answer in English.

[User Question]
```json
{{
    \"instruction\": \"{request_data}\",
}}
```

[The Start of Assistant A's Answer]
```json
{{
    \"instruction\": \"{answerA}\",
}}
```
[The End of Assistant A's Answer]

[The Start of Assistant B's Answer]
```json
{{
    \"instruction\": \"{answerB}\",
}}
```
[The End of Assistant B's Answer]

[Output Format]
```json
{{
    "explanation": str,
    "compare_result": A/B/C
}}
```